\documentclass[a4paper,twoside]{article}

\usepackage{epsfig}
\usepackage{subcaption}
\usepackage{calc}
\usepackage{amssymb}
\usepackage{amstext}
\usepackage{amsmath}
\usepackage{amsthm}
\usepackage{multicol}
\usepackage{pslatex}
\usepackage{apalike}
\usepackage[bottom]{footmisc}

\usepackage{amsfonts}
\usepackage{url}
\usepackage{float}
\usepackage{booktabs}
\usepackage{multirow}
\usepackage{color}
\usepackage{soul}

\usepackage{SCITEPRESS}     % Please add other packages that you may need BEFORE the SCITEPRESS.sty package.

\begin{document}

\title{TeTIm-Eval: a novel curated evaluation data set\\ for comparing text-to-image models}

\author{\authorname{Federico A. Galatolo\sup{1}\orcidAuthor{0000-0001-7193-3754}, Mario G.C.A. Cimino\sup{1}\orcidAuthor{0000-0002-1031-1959} and Edoardo Cogotti\sup{1}}
\affiliation{\sup{1}Department of Information Engineering, University of Pisa, 56122 Pisa, Italy}
\email{federico.galatolo@ing.unipi.it, mario.cimino@unipi.it}
}


\abstract{Evaluating and comparing text-to-image models is a challenging problem. Significant advances in the field have recently been made, piquing interest of various industrial sectors. As a consequence, a gold standard in the field should cover a variety of tasks and application contexts. In this paper a novel evaluation approach is experimented, on the basis of: (i) a curated data set, made by high-quality royalty-free image-text pairs, divided into ten categories; (ii) a quantitative metric, the CLIP-score, (iii) a human evaluation task to distinguish, for a given text, the real and the generated images. The proposed method has been applied to the most recent models, i.e., DALLE2, Latent Diffusion, Stable Diffusion, GLIDE and Craiyon. Early experimental results show that the accuracy of the human judgement is fully coherent with the CLIP-score. The dataset has been made available to the public.}

\onecolumn \maketitle \normalsize \setcounter{footnote}{0} \vfill

\section{\uppercase{Introduction}}

In recent years, the field of generating images from text has seen unprecedented growth, with numerous text-to-image techniques being developed \cite{txt2img_survey}. 
Even if conditional Generative Adversarial Neural Networks were one of the first deep learning-based architectures proposed for the text-to-image task \cite{stackgan} \cite{txt2img_gan} their promising results are generally limited to low-variability data, as their adversarial learning approach does not scale well to modeling complex, multi-modal distributions \cite{gan_largescale}.
Diffusion Models (DMs), which are constructed from a hierarchy of denoising autoencoders, have recently achieved impressive results in image synthesis and beyond, defining the state-of-the-art in class-conditional image synthesis and super-resolution \cite{dm} \cite{dm_superres}.
Furthermore, unconditional DMs can be used for tasks such as inpainting and colorization, as well as stroke-based synthesis \cite{dm_inpaint}. Simple autoregressive transformers, such as the first version of DALLE, on the other hand, have also produced good results in this field \cite{dalle1}.

In addition, the Contrastive Language-Image Pre-Training (CLIP) neural network has recently emerged \cite{clip}. CLIP has been trained on a wide range of image and text pairs. It can be instructed in natural language to predict the most relevant text snippet given an image without directly optimizing for the task.
CLIP embeddings are also robust to picture distribution change, and have been used to achieve state-of-the-art results on vision and language tasks \cite{clip_vlt} as well as image generation from text, when combined with Diffusion Models. 

In this paper we are going to compare the performance of the most recent text-to-image architectures. In particular we are going to consider DALLE-2, Latent Diffusion, Stable Diffusion, GLIDE and craiyon (formerly known as DALL-E mini).

To evaluate the models, we created a new high-quality dataset called TeTIm-Eval (Text To Image Evaluation) composed of 2500 labelled images and 300 text-image pairs divided into ten classes.
The selected models were then used to generate images from the dataset captions, and quantitative evaluation metrics such as the Fréchet Inception Distance (FID) and the CLIP-score were computed with respect to the ground truth images.
Finally, we presented the caption and image from the dataset, as well as a generated image, to human evaluators to assess each model's ability to generate realistic images. 

The remainder of the paper is structured as follows: Section 2 describes the considered models, the proposed dataset, the quantitative evaluation metrics, the design of the human experimentation, and the methods for processing the results. Section 3 illustrates the experimental results, and Section 4 discusses the findings and future research.

\section{\uppercase{Materials and methods}}

\subsection{Generative models}

In this paper we considered five text to image models: DALLE-2, Latent Diffusion, Stable Diffusion, GLIDE and craiyon. The first four are DM based whereas the latter is a decoder-only autoregressive transformer. We excluded the Imagegen model from our study because, while the model description is publicly available \cite{imagegen}, access to the model is not; we also excluded the midjourney model because, while it is accessible online, its implementation is not publicly available.

\subsubsection{Diffusion Models}

Diffusion models have been inspired by non-equilibrium thermodynamics. By first defining a Markov chain of diffusion steps to gradually introduce random noise to data, they are trained to reverse the diffusion process in order to create desired data samples from the noise. Several diffusion-based generative models, such as diffusion probabilistic models \cite{dpm} and denoising diffusion probabilistic models \cite{ddpm}, have been proposed.

Let us define a forward diffusion process for a data point sampled from a real data distribution $x_0 \sim q(x_0)$. In this process we gradually add small amounts of Gaussian noise to the sample as shown in Equation \ref{eq:noise_step} and \ref{eq:noise_process}, resulting in a series of increasingly noisy samples $x_0,\;x_1 \dots\; x_T$. The step sizes are governed by a variance schedule $\beta_t \in (0, 1)$ which can be learned using  the reparameterization trick or held constant as hyperparameter.

\begin{equation}
    q\left(x_t \mid x_{t-1}\right)=\mathcal{N}\left(x_t ; \sqrt{1-\beta_t} x_{t-1}, \beta_t I\right)
    \label{eq:noise_step}
\end{equation}

\begin{equation}
    q\left(x_{1: T} \mid x_0\right)=\prod_{t=1}^T q\left(x_t \mid x_{t-1}\right)
    \label{eq:noise_process}
\end{equation}

The training objective of the diffusion models is, then, to learn the reverse diffusion process $p_\theta(x)$, which is modeled as a Markov chain as shown in Equation \ref{eq:denoise_step} \ref{eq:denoise_process}. This Markov chain that starts with random noise $p_\theta(x_T)=\mathcal{N}(x_T; 0; I)$ and progresses through a succession of less and less noisy samples $x_T,\;x_{T-1} \dots\; x_0$.

\begin{equation}
    p_\theta\left(x_{0: T}\right):=p\left(x_T\right) \prod_{t=1}^T p_\theta\left(x_{t-1} \mid x_t\right)
    \label{eq:denoise_step}
\end{equation}

\begin{equation}
    p_\theta\left(x_{t-1} \mid x_t\right):=\mathcal{N}\left(x_{t-1} ; \mu_\theta\left(x_t, t\right), {\Sigma}_\theta\left(x_t, t\right)\right)
    \label{eq:denoise_process}
\end{equation}

\subsubsection{GPT models for text-to-image}

Generative Pre-trained Transformers are a decoder-only transformer based architecture \cite{transformers} which uses an unidirectional self-attentive model that attends just the tokens before a each token in a sequence \cite{gpt}. GPT models' training objective is then to predict the next token given the previous ones as context.

Despite the fact that GPT models are typically utilized to handle natural language processing tasks, they can be trained to model text and image tokens as a single stream of data in a autoregressive fashion. Using pixels directly as image tokens, on the other hand, would need an excessive amount of memory for high-resolution photos. GPT-based text to image models solved this issue with using two-stage models.

The first stage is to train a discrete variational autoencoder (dVAE) to compress each image into a much smaller grid of image tokens. In the second stage, a GPT transformer is trained on sequences created by concatenating text and image tokens to model the joint distribution over text and images. Finally, to complete the text to image task, the model is given the target text tokens and is used to predict the subsequent image tokens. The predicted image tokens are then fed into the dVAE decoder and projected into the RGB space \cite{dalle1}.

\subsubsection{CLIP}

CLIP is a neural network that was trained on a large set of image and text pairs (400M). CLIP can be used to find the text snippet that best represents a given image, or the most appropriate image given a text query, as a result of this multi-modality training.

In CLIP an image encoder and a text encoder were trained simultaneously to predict the correct pairing of a set of images and text. They were trained to predict, given an image, which one of the randomly sampled text snippets the image was paired to in the training dataset and vice-versa. The model should abstract multiple concepts from the images and from the texts in order to solve the task and the resulting encoders should produce similar embeddings if the image and the text contains similar visual and textual concepts. This approach differs significantly from traditional image tasks, in which the model is typically required to identify a class from a large set of classes (e.g. ImageNet).

More formally give a set of images $x_0, x_1, \dots, x_{n-1}$ and the respective captions $y_0, y_1, \dots, {n-1}$ the features vectors $I_0, I_1, \dots, I_{n-1}$ are computed using the image encoder $I_i = IE(x_i)$ and the features vector $T_0, T_1, \dots, T_{n-1}$ are computed using the text encoder $T_i=TE(y_i)$. Finally the logits cross product is computed as $l = (I \otimes T) \cdot e^\tau$ and the encoder and decoder are trained to solve $n$ joint classification problems (e.g. classify each feature vector $I_i$ with the respective feature vector $T_i$)

\subsubsection{Models taken into account}

In this paper we considered five text to image models: DALLE-2, Latent Diffusion, Stable Diffusion, GLIDE and craiyon.

DALLE-2 architecture \cite{dalle2} is composed by two stages: a prior and a decoder. Given a training dataset of pairs ($x$, $y$) of images $x$ and their corresponding captions $y$, and indicating the CLIP image and text embeddings with $z_i$ and $z_t$, respectively. The prior $P(z_i|y)$ is trained to generate CLIP image embeddings $z_i$ based on captions $y$, and the decoder $D(x|z_i)$ is trained to generate images $x$ based on CLIP image embeddings $z_i$. The prior then learns a generative model of the image embeddings, whereas the decoder inverts images based on their CLIP image embeddings.

For the prior network the continuous vector $z_i$ is directly modelled using a gaussian diffusion model conditioned on the caption $y$.
The decoder $D(x|z_i)$ is also modeled with a gaussian diffusion model as in \cite{glide}, but with four additional context tokens encoding the CLIP embeddings concatenated to the text encoder output. Finally, to obtain high resolution images, the output of the decoder is passed to a two-stage diffusion upsamper model which upsamples the output from 64x64 to 256x256 and from 256x256 to 1024x1024.

Latent Diffusion and Stable Diffusion  \cite{latent-diffusion}  are the same architecture trained on different dataset using a different set of hyperparameters. Latent Diffusion is a two-stage architecture. The first stage is a self-supervised encoder-decoder architecture where the encoder $z = E(x)$ encodes the image $x$ into the embeddings $z$ and the decoder $\hat{x} = D(z)$ decodes the embeddings $z$ back into the original image $\hat{x}$. The second stage is a diffusion process on the latent space $z$ conditioned on image captions $z_T = LDM(z_{T-1} | y)$. The output image is then computed using the decoder on the last step of the diffusion process $D(Z_t)$.

A Vector Quantized Variational AutoEncoder (VQ-VAE) \cite{vqvae} \cite{vq-taming} trained by combining a perceptual loss and a patch-based adversarial objective was used for the first stage. This guarantees that the reconstructions are restricted to the images manifold by imposing local realism and avoids the bluriness induced by relying exclusively on pixel-space losses. A Vector Quantized regularization was also used to avoid high variance latent space. It was used an encoder with a downsample factor of 8.

The second stage is a denoising UNet \cite{unet} conditioned on text embeddings implementing a diffusion process in the latent space $z$. The CLIP ViT-L/14 text encoder is used to compute the text embeddings, and the conditioning is achieved by augmenting the UNet backbone with the cross-attention mechanism. $y$ is encoded with a modality-specific encoder and used to compute keys and values for the attention layers to support conditioning from various modalities (text, semantic maps, etc.); as queries, the flattened internal states of the UNet have been used.

Guided Language to Image Diffusion for Generation and Editing (GLIDE) \cite{glide} is a large diffusion model from OpenaAI. GLIDE implements a text-conditioned diffusion model directly on the pixel space. It implements an Ablated Diffusion Model (ADM) as proposed in \cite{dm} augmented with text conditioning information. Formally for each image of the forward diffusion process $x_t$ and corresponding text caption $y$, GLIDE predicts $DM(x_{t-1} | x_t, y)$. To condition the diffusion model with text the descriptive caption $y$ is encoded using a Transformer and the last embedding of the sequence is used in place of the class conditional token in the ADM architecture; moreover the overall last layer embeddings are projected to the correct dimension and concatenated to the attention context of each ADM layer. Finally each generated images is upsampled from a 64x64 dimension to a 256x256 using a upsample diffusion model.

Craiyon \cite{craiyon} (formerly known as DALL-E mini) is community trained reduced istance of a DALLE \cite{dalle1} model. Craiyon uses a two-stage training method. As in Latent and Stable Diffusion, the first stage is a self-supervised encoder-decoder architecture. Craiyon uses a VQGAN \cite{vq-taming} with a reduction factor of 16 pretrained on the ImageNet dataset for this stage. A GPT-like autoregressive transformer is used in the second stage. A BPE tokenizer is used to encode the text caption $y$, and the target image is encoded in 32x32=1024 image tokens. The text tokens (up to 256) and image tokens are then concatenated and used to train the autoregressive transformer. The overall procedure is equivalent to maximizing the evidence lower bound on the model distribution's joint likelihood over images $x$ and captions $y$.

\subsection{TeTIm-Eval dataset}

\begin{figure}[ht]
    \centering
    \includegraphics[width=\linewidth]{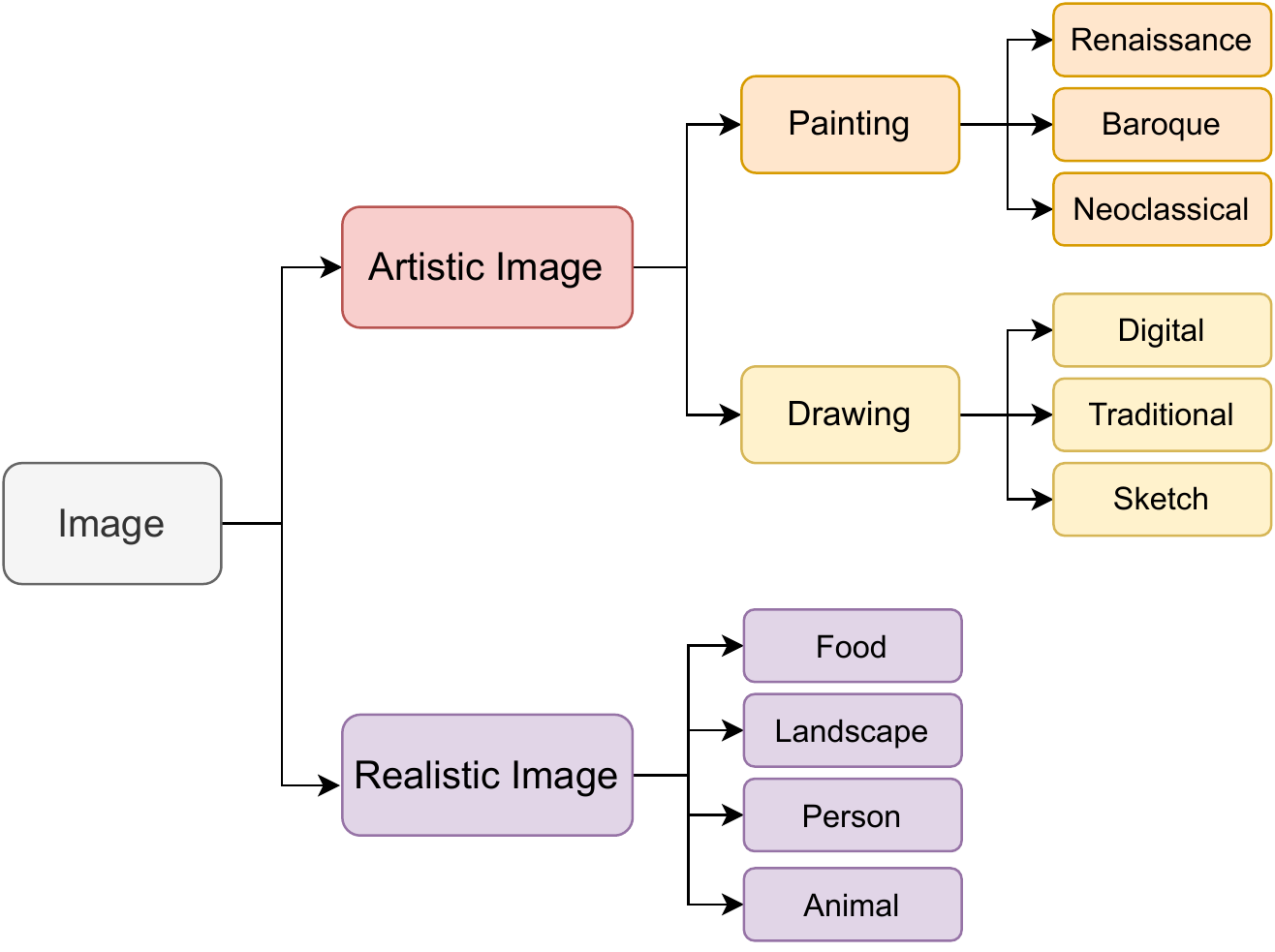}
    \caption{TeTIm-Eval categories and sub-categories taxonomy}
    \label{fig:taxonomy}
\end{figure}

To assess the performance of the models under evaluation, we created a diverse and high-quality dataset of text and image pairs that we called TeTIm-Eval (Text To Image Evaluation).
Existing datasets and category-specific websites were used to create the dataset.
We considered three types of images: paintings, drawings, and realistic photographs. And, as shown in Figure \ref{fig:taxonomy}, we identified the following sub-categories for each category.

\begin{itemize}
    \item Painting
    \begin{itemize}
        \item Renaissance paintings
        \item Baroque paintings
        \item Neoclassical paintings
    \end{itemize}
    \item Drawing
    \begin{itemize}
        \item Digital Drawings
        \item Traditional Drawings
        \item Sketch Drawings
    \end{itemize}
    \item Realistic Photos
    \begin{itemize}
        \item Food Photos
        \item Landscape Photos
        \item Person Photos
        \item Animal Photos
    \end{itemize}
\end{itemize}

We first identified the sources and downloaded the available data, then sampled the downloaded data at random and manually rejected images that did not meet our quality criteria, yielding a total of 2500 images (250 per sub-category).
Finally, in order to create the final dataset, we randomly selected 30 images from each sub-category and manually wrote a textual description for each of the 300 images.

We followed these quality criteria to ensure the creation of an high-quality dataset:

\begin{itemize}
    \item Only images that clearly belong to the sub-category
    \item Only images where the content fills the available space (no frames, etc.)
    \item Only images released under the Creative Commons License
    \item Only high definition images
    \item No signatures or watermarks
    \item No adult-rated, violent or hate images
    \item No political images
\end{itemize}

\begin{table}[ht]
\centering
\small
\begin{tabular}{@{}lll@{}}
\toprule
Category & Sub-category & Source \\
\midrule
\multirow{3}{*}{Painting}  & Renaissance   & Wikiart \\
                           & Baroque       & Wikiart \\ 
                           & Neoclassical  & Wikiart \\
\midrule
\multirow{3}{*}{Drawing}   & Digital       & Deviantart and Openverse \\
                           & Traditional   & Deviantart and Openverse \\
                           & Sketch        & ImageNet Sketch \\
\midrule
\multirow{4}{*}{Realistic} & Food          & Wikimedia Commons \\
                           & Landscape     & Wikimedia Commons \\
                           & Person        & COCO \\
                           & Animal        & Wikimedia Commons \\
\bottomrule \\
\end{tabular}
\caption{Dataset sources}
\label{table:dataset-sources}
\end{table}

We selected a total of 6 different data sources: Wikiart \cite{wikiart} which is an online, user-editable visual art encyclopedia, Deviantart \cite{deviantart} which is an online art community, Openverse \cite{openverse} which is an open-source search engine for open content developed as part of the WordPress project, ImageNet Sketch \cite{imagenet-sketch} which is a dataset consisting of 50000 images (50 images for each of the 1000 ImageNet\cite{imagenet} classes), Wikimedia Commons \cite{wikimedia} which is a media repository of open images, sounds, videos and other media from the Wikimedia Foundation and COCO \cite{coco} which is a large-scale object detection, segmentation, and captioning dataset. All the painting images where sampled from Wikiart, digital and traditional drawings from Deviantart and Openverse, the sketches from ImageNet Sketch, the food landscape and animal photos from Wikimedia Commons and finally the person photos from the COCO dataset as shown in Table \ref{table:dataset-sources}. Finally the captions were written by the same person to ensure consistency across the whole dataset.
Furthermore given the dataset's curated and high-quality nature, it can also be used to train and/or evaluate zero/few shot learning algorithms, which are notorious for requiring high-quality data to perform well. 

The overall dataset, the 2500 labelled images and the 300 text-image pairs, as well as its companion source code has been released on GitHub \cite{tetim-eval-repo}.

\section{\uppercase{Experimental results and discussion}}

We compared the models taken into account using 300 image-text pairs and three evaluation metrics: the CLIP score, the Fréchet Inception Distance (FID), and the human evaluation.
Specifically, firstly, using the CLIP text encoder, we computed the feature vector of each target text $TE(y)$. Then, using the CLIP image encoder $IE(x_i)$, we computed the CLIP score as its dot product with the feature vectors of the generated images. Formally given a target text $y$ and a set of generated images $x_0, x_1, \dots, x_{n-1}$ the CLIP score of the image $x_i$ with respect to the target text $y$ is $CS_i = TE(y) \cdot IE(x_i)$. Even if the CLIP score is a good metric to determine the semantic distance between captions and images it is important to point out that some of the models under study directly use CLIP as part of their pipeline like DALLE2 or use CLIP indirectly to filter the training dataset (like Stable and Latent Diffusion)

The Fréchet Inception Distance(FID) is a widely used to assess the quality of images created by a generative models\cite{fid_gan}. The FID is the Multivariate Gaussian Fréchet distance of the probability distributions of the features extracted using an Inception V3 \cite{inception-v3} model pre-trained on the ImageNet dataset. Formally given the probability distribution of the inception features extracted from the real images $\mathcal{N}(\mu_r, \Sigma_r)$ and the one of the inception features extracted from the generated images $\mathcal{N}(\mu_g, \Sigma_g)$ the FID can be computed as $FID = |\mu_r - \mu_g| + tr(\Sigma_r + \Sigma_g - 2(\Sigma_r\Sigma_g)^{\frac{1}{2}})$. Even if the FID is the most used metric to compare text to image models it has been shown \cite{fid_bias} that for a low number of samples it is not reliable, because it frequently fails to reflect the goodness and fidelity of generated images. For this reason it is not used in the current experimentation. 
For the Human evaluation we developed a web platform in which each user was prompted with a descriptive caption and two images: the real image from the TeTIm-Eval dataset and an image generated by one of the models under study. We then asked the user to identify the real image.

\begin{table}[ht]
\small
\centering
\begin{tabular}{@{}lc@{}}
\toprule
    Title & Value   \\
\midrule
    Participants & 183 \\
    Answers    & 5010 \\
    Right Answers & 3419 \\
\bottomrule \\
\end{tabular}
\caption{Overall human evaluation statistics}
\label{table:human-stats}
\end{table}

\begin{table}[ht]
\small
\centering
\begin{tabular}{@{}lc@{}}
\toprule
    Metric & Value   \\
\midrule
    Accuracy            & .68 \\
    False Positive Rate & .34 \\
    False Negative Rate & .30 \\
    Precision           & .67 \\
    Recall              & .70 \\
\bottomrule \\
\end{tabular}
\caption{Overall human evaluation performance}
\label{table:human-overall-performance}
\end{table}

Table \ref{table:human-stats} and table \ref{table:human-overall-performance} show the human involved and their answers, as well as their overall performance, respectively. Table \ref{table:human-performance-category} shows the human performance per category. Finally Table \ref{table:overall-performance-model} displays the human performance per model as well as its CLIP-score, showing that human accuracy is consistent with the CLIP score. It is important to note that in this context, accuracy is defined as the number of correctly identified human-generated images divided by the total number of answers. As a result, the lower the accuracy, the better the model is at producing images that fool humans into thinking they were created by humans. In this context, the lower the human accuracy, the better the model. 

\begin{table}[ht]
\small
\centering
\begin{tabular}{@{}llccc@{}}
\toprule
Category & Sub-Category & H. Acc. & H. Prec. & H. Rec.  \\
\midrule
\multirow{3}{*}{Painting}  & Renaissance    & .73 & .73 & .74 \\
                           & Baroque        & .77 & .74 & .80 \\
                           & Neoclassical   & .73 & .69 & .79 \\
\midrule
\multirow{3}{*}{Drawing}   & Digital        & .63 & .68 & .49 \\
                           & Traditional    & .60 & .63 & .56 \\
                           & Sketch         & .62 & .58 & .66 \\
\midrule
\multirow{4}{*}{Realistic} & Food           & .66 & .62 & .77 \\
                           & Landscape      & .65 & .61 & .70 \\
                           & Person         & .77 & .76 & .80 \\
                           & Animal         & .68 & .66 & .74 \\
\bottomrule \\
\end{tabular}
\caption{Overall human performance per category}
\label{table:human-performance-category}
\end{table}

\begin{table}[ht]
\small
\begin{tabular}{lcc}
\toprule
Model &  H. Accuracy ($\downarrow$)  &  CLIP Score ($\uparrow$) \\
\midrule
Stable Diffusion &      .62 &     29 \\
DALLE2           &      .63 &     29 \\
Craiyon          &      .71 &     28 \\
Latent Diffusion &      .71 &     26 \\
GLIDE            &      .73 &     22 \\
\bottomrule
\end{tabular}
\caption{Overall human performance per model}
\label{table:overall-performance-model}
\end{table}

\section{\uppercase{Conclusions}}

In this paper, a novel evaluation method for text-to-image models is introduced. A novel high-quality evaluation dataset consisting of 2500 labelled images and 300 text-image pairs has been released to the public. Five different state-of-the art models have been evaluated, both using a quantitative metric and humans involving about 180 participants, for 5K+ answers.
The human evaluation shows that Stable Diffusion is the most accurate model, followed by DALLE2, Craiyon, Latent Diffusion and GLIDE. The CLIP-score metric is coherent with the human evaluation. Future work will consider a larger data set and will include the FID metrics. Given the high quality of the dataset, it is worth noting that it can also be used to train and/or evaluate zero/few shot learning algorithms. 

\section*{\uppercase{Acknowledgements}}
This work has been partially supported by:
(i) the National Center for Sustainable Mobility MOST/Spoke10, funded by the Italian Ministry of University and Research in the framework of the National Recovery and Resilience Plan; (ii) the PRA\_2022\_101 project “Decision Support Systems for territorial networks for managing ecosystem services”, funded by the University of Pisa; (iii)
the Ministry of University and Research (MUR) as part of the PON 2014-2020 “Research and Innovation" resources – Green/Innovation Action – DM MUR 1061/2022";
the MUR, in the framework of the FISR 2019 Programme, under Grant No. 03602 of the project “SERICA”; the Tuscany Region in the framework of the SecureB2C project, POR FESR 2014-2020, Project number 7429 31.05.2017;
the MUR, in the framework of the "Reasoning" project, PRIN 2020 LS Programme, Project number 2493 04-11-2021.

\bibliographystyle{apalike}
{
\small
\bibliography{references}
}

\end{document}